\documentclass[sn-mathphys-num]{sn-jnl}% Math and Physical Sciences Numbered Reference Style 
\usepackage{graphicx}%
\usepackage{multirow}%
\usepackage{amsmath,amssymb,amsfonts}%
\usepackage{amsthm}%
\usepackage{euscript}%
\usepackage[title]{appendix}%
\usepackage{xcolor}%
\usepackage{textcomp}%
\usepackage{manyfoot}%
\usepackage{booktabs}%
\usepackage{algpseudocode}%
\usepackage{listings}%
\usepackage{lmodern}
\usepackage[utf8]{inputenc} % Para codificación UTF-8
\usepackage[T1]{fontenc}    % Para una codificación de salida correcta
\usepackage{pgf-pie}
\usepackage{pgfplots} 
\pgfplotsset{compat=1.18} % Configuración de compatibilidad
\usepackage{pgfplotstable}
\usepackage{tikz}
\usepackage{caption}
\usepackage{subcaption}

\begin{document}
\UseRawInputEncoding

\title[Article Title]{Automated Detection of Clinical Entities in Lung and Breast Cancer Reports Using NLP Techniques}


\author*[a]{\fnm{J.} \sur{Moreno-Casanova}} \email{julia.moreno.casanova@gmv.com}

\author[a]{\fnm{J.M.} \sur{Au\~{n}\'on}}

\author[b]{\fnm{A.} \sur{M\'artinez-P\'erez}}

\author[b]{\fnm{M.E.} \sur{P\'erez-Mart\'inez}}
\author[b]{\fnm{M.E.} \sur{Gas-L\'opez}}

\affil[a]{\orgdiv{Department of Artificial Intelligence and Big Data}, \orgname{GMV}, \orgaddress{\street{Isaac Newton 11}, \city{Tres Cantos}, \postcode{28760}, \state{Madrid}, \country{Spain}}}

\affil[b]{\orgname{Health Research Institute Hospital La Fe}, \orgaddress{\street{Avinguda de Fernando Abril Martorell, 106}, \city{Valencia}, \postcode{46026}, \state{Valencia}, \country{Spain}}}


\abstract{Many clinical research projects, including those focused on cancer, rely on the manual extraction of information from clinical reports. This process is both time-consuming and prone to errors, limiting the efficiency of data-driven approaches in healthcare. To address these challenges, Natural Language Processing (NLP) offers an alternative for automating the extraction of relevant data from electronic health records (EHRs). In this study, we focus on lung and breast cancer due to their high incidence and the significant impact they have on public health. Early detection and effective data management in both types of cancer are crucial for improving patient outcomes.

To enhance the accuracy and efficiency of data extraction, we utilized GMV's NLP tool uQuery, which excels at identifying relevant entities in clinical texts and converting them into standardized formats such as SNOMED and OMOP. uQuery not only detects and classifies entities but also associates them with contextual information, including negated entities, temporal aspects, and patient-related details. 

In this work, we explore the use of NLP techniques, specifically Named Entity Recognition (NER), to automatically identify and extract key clinical information from EHRs related to these two cancers. A dataset from Health Research Institute Hospital La Fe (IIS La Fe), comprising 200 annotated breast cancer and 400 lung cancer reports, was used, with eight clinical entities manually labeled using the Doccano platform. To perform NER, we fine-tuned the bsc-bio-ehr-en3 model, a RoBERTa-based biomedical linguistic model pre-trained in Spanish. Fine-tuning was performed using the Transformers architecture, enabling accurate recognition of clinical entities in these cancer types. Our results demonstrate strong overall performance, particularly in identifying entities like MET and PAT, although challenges remain with less frequent entities like EVOL. The incorporation of a text pre-processing layer significantly improved the accuracy of entity recognition, reflected in high precision, recall, and F1 scores across most entities in both validation and test sets.
}

\maketitle

\section{Introduction}\label{sec1}

As is well known, lung cancer is one of the most common and deadly carcinomas worldwide \cite{bib1}. In fact, its survival rate is lower compared to many other major cancers \cite{bib1}. Thus, in order to improve the patient's survival rate, early detection is key. However, lung cancer is not the only carcinoma that stands out in this aspect. Although breast cancer has a higher survival rate than lung cancer, in 2020 2.3 million women were diagnosed with it, with 685,000 deaths reported worldwide  \cite{bib2}. So, as with lung cancer, early and accurate detection can save many lives.

Given the critical importance of early detection, it becomes imperative to leverage comprehensive patient data to inform clinical decisions. This data is predominantly stored in electronic health records (EHRs), which have become integral to modern clinical care. Among the various types of data in EHRs, clinical notes are particularly valuable as they contain detailed, narrative descriptions of patient conditions, treatments, and responses. These records present a significant opportunity for generating large-scale real-world evidence (RWE) that can inform the development of new biomarkers, therapies, and clinical decision-support systems. However, extracting actionable insights from EHRs is challenging due to the prevalence of unstructured data formats, such as free-text notes, which are often complicated by specialized medical terminology, abbreviations, and unique documentation practices.
 
To address these challenges, researchers have developed various natural language processing (NLP) techniques for automating the clinical information extraction process. Clinical named entity recognition (NER) is a critical NLP task that focuses on identifying and categorizing key clinical entities, such as medical conditions, treatments, and tests. By automating this process, the time and effort required for manual chart review and coding by health professionals can be significantly reduced, thus improving patient care efficiency and accelerating clinical research.

It should be noted that this methodology is based on GMV's NLP tool \texttt{uQuery}\cite{bib9}, which excels at identifying relevant entities in clinical texts and converting them into standardized formats such as SNOMED and OMOP. \texttt{uQuery} not only detects and classifies entities but also associates them with contextual information, including negated entities, temporal aspects, and patient-related details. Additionally, the tool integrates with \texttt{doccano} \cite{doccano}, an open-source correction system, allowing for manual adjustments and enhancements to the entity recognition process. This ensures that entities are accurately captured and refined in future iterations. \texttt{uQuery} also features a significant pre-processing layer that effectively captures a wide range of entities, minimizing the risk of losing essential information.

The structure of this document is as follows: Section 2 provides an overview of the study's background, objectives, literature review, and the significance or contribution of this study to the field. Section 3 describes the methodology, including dataset details, model training, and validation procedures. Section 4 presents the results and statistical analyses, and Section 5 concludes with a summary of findings and their implications for cancer research.

\section{Background}\label{sec2} 

In this section, various algorithms that provide the background for the methods proposed in this paper are reviewed, as introduced in  \ref{sec1}. These algorithms are essential for understanding the theoretical and practical aspects of the approach, offering a comprehensive background to the innovations and improvements presented.

First of all, the work of Fang et al. \cite{bib5} will be reviewed. This study focuses on extracting clinical entities from Chinese electronic medical records (EMRs) of patients with pituitary adenomas (a common pituitary disorder affecting young adults). The goal is to enable machines to intelligently process and automatically extract clinical named entities from unstructured texts in EMRs.Data from a neurosurgery treatment center in China were used, analyzing 500 patient records \cite{bib5}. Four methods were applied: dictionary-based matching, Conditional Random Fields (CRF), BiLSTM-CRF, and BERT-BiLSTM-CRF. Specifically, the BERT-BiLSTM-CRF model achieved the highest performance in most cases, particularly in extracting symptoms, body regions, and diseases \cite{bib5}. The study demonstrates that deep learning methods are effective in extracting clinical entities from Chinese electronic medical records (EMRs), highlighting their potential for broader application in diverse medical texts. 

This research aligns with our objective of developing methods for the effective and accurate extraction of clinical information. Furthermore, our study aims to address the limitations identified in Fang et al.'s work by applying these techniques to a different dataset, specifically focusing on lung and breast cancer reports, and enhancing performance through further refinements tailored to these contexts.

Several researchers have adopted similar approaches: Zhang et al. \cite{bib10} used a fine-tuned BERT model to extract clinical information specific to breast cancer-related texts, demonstrating the model's adaptability to different types of clinical data. Obeid et al. \cite{bib11} utilized a convolutional neural network (CNN) to detect mental status indicators within emergency department clinical notes, demonstrating the versatility of machine learning models in various clinical contexts. These studies underscore the significance of leveraging advanced machine learning models for clinical named entity recognition (NER), which is a critical component of our proposed methodology. Our work builds on these approaches by integrating additional contextual information and refining the models to enhance accuracy and applicability.

Another interesting paper is that of Paolo D, Bria A, Greco C, et al.\cite{bib6}. This study addresses NER for extracting clinical information from Italian EHRs of patients with non-small cell lung cancer (NSCLC). The research focuses on improving a previous model by expanding the set of clinical entities from 25 to 29, including negated entities (e.g., absence of symptoms). The authors applied a BERT-based model \cite{bib6} to an annotated dataset of 257 NSCLC patients (CLARO \cite{bib6})  and evaluated its performance. Results showed that including negated entities increased the complexity and slightly decreased the model's performance. Despite this, the study highlights the effectiveness of NER in extracting comprehensive clinical information (in our paper, we will demonstrate similar effectiveness in applying this technique to our specific dataset), including both present and absent health characteristics, which could enhance patient care and support personalized health and prognostic tasks \cite{bib6}.

These studies underscore the significance of advanced algorithms in the accurate extraction and analysis of clinical information, providing a foundation for the methodologies proposed in this paper.

\section{Methods}\label{sec3} 
\begin{figure}[h!]
    \centering
    \includegraphics[width=0.9\textwidth]{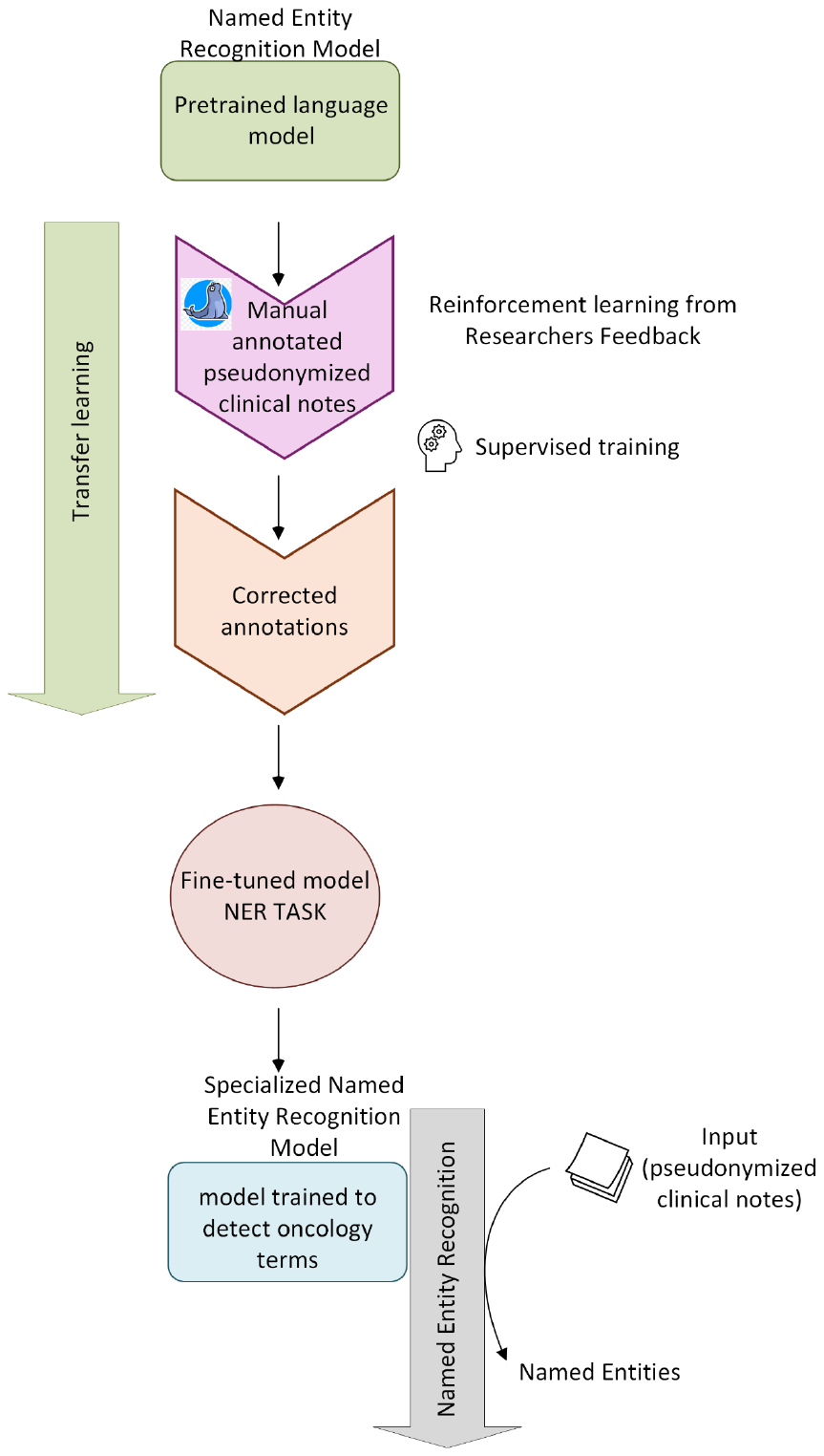}
    \caption{Proposed method pipeline, illustrating the stages of the process, from dataset preparation, fine-tuning of the cancer-specific NER model, to the application of the model to pseudonymized clinical records extracted from the Big Data Platform of IIS La Fe and the detection of entities within them.}\label{fig1}
\end{figure}

In this chapter, the methodologies and processes followed to apply NER to clinical texts will be explained. The dataset composition is first described, followed by an explanation of the selected model and the techniques used. A method pipeline, as illustrated  in Figure \ref{fig1}  integrates all stages of the process, starting with data preparation. This is followed by the fine-tuning of a cancer-specific NER model and its subsequent application to pseudonymized clinical records extracted from the IIS La Fe Big Data Platform, culminating in the detection of relevant entities, such as EVOL (evolution), FACTR (risk factors), ANTPERSON (personal history, specific to lung cancer), MUTAC (genetic mutations, specific to lung cancer), MET (method of diagnosis), PAT (pathology), SINT (symptomatology), and TTO (treatment). This integrated approach ensures an effective implementation of NER, enabling the extraction of clinically relevant information while adhering to established guidelines and regulations.

\bmhead{Ethical Approval and Informed Consent Statement}
All procedures in this study adhered to established guidelines and regulations. The study received approval from the relevant legal and ethical boards on 01 26, 2024. This included the ethics committee for biomedical research involving medicines at the University and Polytechnic La Fe Hospital in Valencia, with registration number  2021-686-1. This committee adheres to Good Clinical Practice (GCP) standards (CPMP/ICH/135/95) and complies with the applicable legal frameworks governing its operations. The committee confirmed that there were no conflicts of interest in the assessment and approval of the study, affirming that the project aligns with ethical guidelines for biomedical research involving human subjects and is feasible in terms of its scientific approach, objectives, materials, and methods as detailed in the application.

Due to retrospective nature of the study the informed consent has been waived by the approval committee. The legal basis for processing personal data is supported by anonymized or pseudonymized data handling, as permitted under Spanish law, particularly Article 16.3 of Law 41/2002 of November 14, which governs patient rights and clinical information, and in the second paragraph of the seventeenth additional provision regarding health data processing in Organic Law 3/2018 of December 5, on the Protection of Personal Data and the Guarantee of Digital Rights.

\subsection{Data Source}\label{subsec1}
The University and Polytechnic La Fe Hospital in Valencia serves as the main clinical facility within the La Fe Health Department. This health district, which supports a population of approximately 300,000, includes two specialized centers and twenty primary care facilities. The Hospital's Electronic Health Record (EHR) system integrates data from both primary and specialized care services.
The La Fe Health Department has implemented an EHR system across various care levels, accumulating over 20 million records. This system has achieved stage 6 in the eight-stage EMRAM maturity model, which ranges from stage 0 to stage 7. Presently, the data lake consists of both structured and semi-structured data from multiple information systems involved in clinical activities such as emergency care, outpatient services, hospital admissions, clinical reports, surgical units, intensive care, and home-based hospital care. Additionally, the La Fe Health Department has developed a Real-World Data analysis platform that aggregates data from 22 datamarts, encompassing 750 million rows, 84 tables, and 4,064 columns.

This study is a retrospective, observational analysis conducted at a single center. It includes all adult individuals diagnosed with breast or lung cancer 

For the breast cancer cohort, patients aged over 18 at the time of diagnosis and belonging to the hospital health department (DS7LAFE) registered in primary care and outpatient clinics with ICD-9 and ICD-10 codes referring to breast cancer were included. Oncology, radiotherapy, thoracic surgery, internal medicine and breast tumour committee services were filtered for the extraction of pathological anatomy notes. A similar approach was followed for the lung cancer cohort as for the breast cancer cohort. ICD codes for lung cancer were identified and patients were selected who were 18 years of age or older and belonged to the hospital. In particular, clinical pathology notes from the oncology, radiotherapy, thoracic surgery, internal medicine and pneumology departments were selected. For the symptomatology notes, a sample of 2000 random reports was selected from the medical records of all patients in the lung cancer cohort.

Once the cohorts were defined, the dataset that GMV would work with was extracted. It consists on 6000 unstructured reports, of which 2000 were clinical notes on breast cancer pathology and 4000 were clinical notes on lung cancer, of which 2000 were related to pathology and another 2000 to symptomatology. From this dataset, it was agreed with GMV that IIS La Fe would only tag 600 reports for the model learning (200 from each of the indicated categories).
 
In both cases, information was collected from anatomical pathology reports, randomly selecting 2000 anonymised clinical notes from breast cancer patients and 2000 from lung cancer patients. In addition, for lung cancer, additional information on symptomatology was extracted from historical hospital records, obtaining a random sample of 2000 additional notes. Once the study dataset was obtained, access was granted to GMV through a secure mechanism, guaranteeing the security, integrity and confidentiality of the data at all times. GMV selected randomly 200 notes from each category and uploaded them to Doccano for carring out the tagging task.

\subsection{Data Preparation}\label{subsec2}

Before applying the Named Entity Recognition (NER) model, a fine-tuning process was conducted to adapt and optimize the pre-trained model for the specific NER task. This process is nothing more than a method commonly used to adjust and improve a pre-trained machine learning model for a specific task (in this case NER task) using a specific dataset. The dataset, consisting of 200 breast cancer reports (pathological anatomy) and 400 lung cancer reports (200 pathological anatomy and 200 symptomatology), was utilized for this purpose. The dataset included annotations for eight distinct categories: six common to both cancer types (EVOL - Evolution, FACTR - Risk Factors, MET - Method of Diagnosis, PAT - Pathology, SINT - Symptomatology, and TTO - Treatment) and two specific to lung cancer (ANTPERSON - Personal history of malignant neoplasm and MUTAC - mutations associated with lung cancer).
To facilitate the NER task, a \textit{dictionary} of relevant terms for each entity was created. These terms were collected from databases, literature, pre-existing lists, and ontologies, and served as the foundation for the model's recognition of key entities in the text corpus.It is crucial to ensure that the \textit{dictionary }is broad enough to capture relevant entities but not too general, as it could include irrelevant terms. The resulting \textit{dictionary}, developed by researchers at IIS La Fe and professionals at Hospital La Fe, was structured hierarchically, where entities are grouped by categories, facilitating their integration into text processing pipelines. This hierarchy consisted of the following categories: EVOL (Evolution), FACTR (Risk Factors), ANTPERSON (Personal History), MUTAC (Genetic Mutations), MET (Diagnostic Method), PAT (Pathology), SINT (Symptomatology), and TTO (Treatment). The creation of the \textit{dictionary} involves a structured process of collecting, cleaning, structuring, and normalizing terms, essential for the model to learn to recognize specific entities in a text corpus. After creating the \textit{dictionary}, the text annotation was carried out using the Doccano \cite{doccano} tool. 

Doccano is an open-source platform designed for data annotation and labeling in natural language processing (NLP) and machine learning projects (as mentioned in section \ref{sec1})). To carry out this process, GMV provided a secure environment for IIS La Fe researchers to manually annotate the texts. The annotation consisted of highlighting text fragments associated with one of the defined labels. To ensure high-quality annotation, iterative annotation cycles were performed, allowing for reviews and adjustments. Once the annotated corpus was available, the data was stored in the appropriate format to feed the machine learning model.
\vspace{3cm}

\subsection{Modelling}\label{subsec3}

Following on from the previous subsection, once the data has been prepared, we proceed to fine-tune the model so that it can perform the task of identifying entities. The model used for this task is bsc-bio-ehr-en \cite{bib7}, a biomedical linguistic model based on the RoBERTa (Robustly Optimized BERT Pretraining Approach) architecture \cite{bib3}, which is pre-trained in Spanish and suitable for medical records and EHRs. Thus, the fine-tuning process adapted the model for NER, specifically designed to detect relevant entities in breast and lung cancer clinical reports. A detailed explanation of the model's application is provided in the following subsection

\subsubsection{Model use}\label{subsubsec1}

The bsc-bio-ehr-en3 model, based on the widely adopted Transformer architecture, was employed to perform Named Entity Recognition (NER) tasks. Transformers have had a transformative impact on natural language processing (NLP), achieving remarkable success in tasks such as machine translation, text generation, and sentiment analysis. Building on this powerful architecture, the bsc-bio-ehr-en3 model was fine-tuned specifically to detect relevant entities in clinical reports. The following section outlines the implementation of this model for entity detection in the context of breast and lung cancer reports:

\vspace{0.5cm}
\definecolor{codegreen}{rgb}{0,0.6,0}
\definecolor{codegray}{rgb}{0.5,0.5,0.5}
\definecolor{codepurple}{rgb}{0.58,0,0.82}
\definecolor{backcolour}{rgb}{0.95,0.95,0.92}

\lstdefinestyle{mystyle}{
    backgroundcolor=\color{backcolour},   
    commentstyle=\color{codegreen},
    keywordstyle=\color{magenta},
    numberstyle=\tiny\color{codegray},
    stringstyle=\color{codepurple},
    basicstyle=\ttfamily\footnotesize,
    breakatwhitespace=false,         
    breaklines=true,                 
    captionpos=b,                    
    keepspaces=true,                 
    numbers=left,                    
    numbersep=5pt,                  
    showspaces=false,                
    showstringspaces=false,
    showtabs=false,                  
    tabsize=2
}

\lstset{style=mystyle}

\begin{lstlisting}[language=python]
from transformers import AutoTokenizer, AutoModelForTokenClassification, pipeline

tokenizer = AutoTokenizer.from_pretrained("/modelo_cmama_cpulmon", local_files_only=True)

model = AutoModelForTokenClassification.from_pretrained("/modelo_cmama_cpulmon", local_files_only=True)

pipe = pipeline('ner', tokenizer=tokenizer, model=model, aggregation_strategy="average")

sentence = 'Por el hallazgo de múltiples fracturas por estrés, se procedió a estudio en nuestras consultas, realizándose análisis con función renal, calcio sérico y urinario, calcio iónico, magnesio y PTH, que fueron normales.'
results = pipe(sentence)

print(results)
\end{lstlisting}

\section{Results and Discussion}\label{sec4} 
This section outlines the results obtained from the evaluation of the model with both the breast cancer report set and the combined dataset (comprising breast and lung cancer reports set). Once the results have been presented, they will be analyzed and discussed.
\vspace{3cm}
\subsection{Validation results for the breast cancer reporting set}\label{subsec4}

To initially assess the model's performance, a subset of 200 breast cancer reports (pathological anatomy) was utilized. This set was randomly split according to these percent- ages: 50 \% train, 25 \% test and 25 \% validation. Thus, the distribution of the number of labels in the set would be as follows:

\begin{table}[h]
\caption{Label Distribution in the Breast Cancer Set}\label{tab1}%
\begin{tabular}{@{}lllllll@{}}
\toprule
Set & EVOL & FACTR  & MET & PAT & SINT & TTO\\
\midrule
Train    & 0   & 48  & 1237  & 814   & 54   & 104   \\
Validation    & 2  & 19   & 647   & 410  & 16  & 53  \\
Test    & 4 & 22  & 759  & 544  & 38  & 89  \\
Complete   & 2  & 19  & 647  & 410  & 16  & 53  \\
\botrule
\end{tabular}
\end{table}

As shown in Table \ref{tab1}, the most frequently annotated entities categories in the breast cancer dataset are MET (Method of Diagnosis) and PAT (Pathology). These two entities dominate the label distribution across the training, validation, and test sets. On the other hand, FACTR (Risk Factors) and EVOL (Evolution) are considerably less represented, with EVOL being absent in the training set.
The underrepresentation of certain entities, such as EVOL, poses a significant challenge for the model's performance, particularly in terms of generalizability. Since there are no instances of EVOL in the training set, the model will struggle to detect or predict this entity during testing, leading to either null or erroneous results. Similarly, entities with sparse representation (like FACTR and SINT) may also suffer from limited predictive accuracy.

\subsubsection{Automatic validation}\label{subsubsec2}

This section presents the results of the automatic validation conducted during the model training phase. The validation process includes evaluating the model's performance using several key metrics for each label: precision, recall, and F1-score \cite{bib8}. These metrics are essential for assessing the model's effectiveness in entity detection tasks. Additionally, global metrics such as accuracy and loss \cite{bib8} are included to provide a comprehensive view of the model's overall performance.
The following table summarizes the metrics for each label in the breast cancer dataset:
\begin{table}[h]
\caption{Automatic Validation Results for the Breast Cancer Model by Label}\label{tab2}%
\begin{tabular}{@{}lllllll@{}}
\toprule
Validation Set & EVOL & FACTR  & MET & PAT & SINT & TTO\\
\midrule
F1   & 0  & 0.5   & 0.8289   & 0.6578  & 0.7999  & 0.7222  \\
Precision  & 0 & 0.4762  & 0.8252  & 0.6538  & 0.7368  & 0.6964 \\
Recall & 0  & 0.5263  & 0.8288  & 0.6618  & 0.8750  & 0.75  \\
\botrule
\end{tabular}
\end{table}

\begin{table}[h]
\caption{Automatic Validation Results for the Breast Cancer Model }\label{tab3}%
\begin{tabular}{@{}llllll@{}}
\toprule
Validation Set & Accuracy & F1 & Precision & Recall & Loss \\
\midrule
Global  & 0.9418  & 0.7684  & 0.7641   & 0.7728 & 0.2497  \\
\botrule
\end{tabular}
\end{table}

As shown in Table \ref{tab2}, the EVOL label has zero metrics, as previously mentioned, underscoring a significant performance issue for this entity. In contrast, the MET label achieved the highest scores across all metrics, indicating strong performance. Both FACTR and PAT labels show lower metrics compared to others, suggesting areas that require improvement.
Despite these challenges, the overall validation metrics, as presented in Table \ref{tab3}, indicate strong performance. The model achieved an accuracy of  94\%, an F1-score of 0.77, and a low loss of 0.25, reflecting the model's reliability in entity recognition. These results demonstrate that, while there are areas for improvement, the model performs well overall.

The model's generalizability was further assessed using a test set, to ensure that the results from the validation phase hold true across different datasets. The metrics from the test set, shown in Table \ref{tab4}, reveal that the model maintains strong performance in identifying entities, especially for the FACTR, MET, and SINT labels. However, the EVOL label still shows zero metrics, indicating persistent challenges in recognizing this entity.

The overall results for the test set, as presented in Table \ref{tab5}, confirm that the model performs exceptionally well. With an accuracy of 94\% and an F1-score of 0.75, the model demonstrates strong generalization to unseen data. The loss value of 0.24 further supports the conclusion that the model is robust and not overfitting. These results provide confidence in the model's ability to reliably identify entities in real-world clinical reports, although further refinement is needed for certain labels like EVOL.

\begin{table}[h]
\caption{Automatic Test Results for the Breast Cancer Model by Label}\label{tab4}%
\begin{tabular}{@{}lllllll@{}}
\toprule
Test Set & EVOL & FACTR  & MET & PAT & SINT & TTO\\
\midrule
F1   & 0  & 0.7059  & 0.8057   & 0.6503  & 0.6383  & 0.7027  \\
Precision  & 0 & 0.6207  & 0.7747  & 0.6444  & 0.5357 & 0.6771 \\
Recall & 0  & 0.8181  & 0.8392  & 0.6563  & 0.7895  & 0.7303  \\
												
\botrule
\end{tabular}
\end{table}

\begin{table}[h]
\caption{Automatic test results of the breast cancer model}\label{tab5}%
\begin{tabular}{@{}llllll@{}}
\toprule
Test Set & Accuracy & F1 & Precision & Recall & Loss \\
\midrule
Global  & 0.9442  & 0.7464  & 0.7215 & 0.7733 & 0.2376  \\
\botrule
\end{tabular}
\end{table}

\subsubsection{Manual validation}\label{subsubsec3}

\begin{figure}[h]
\centering
\includegraphics[width=1.1\textwidth]{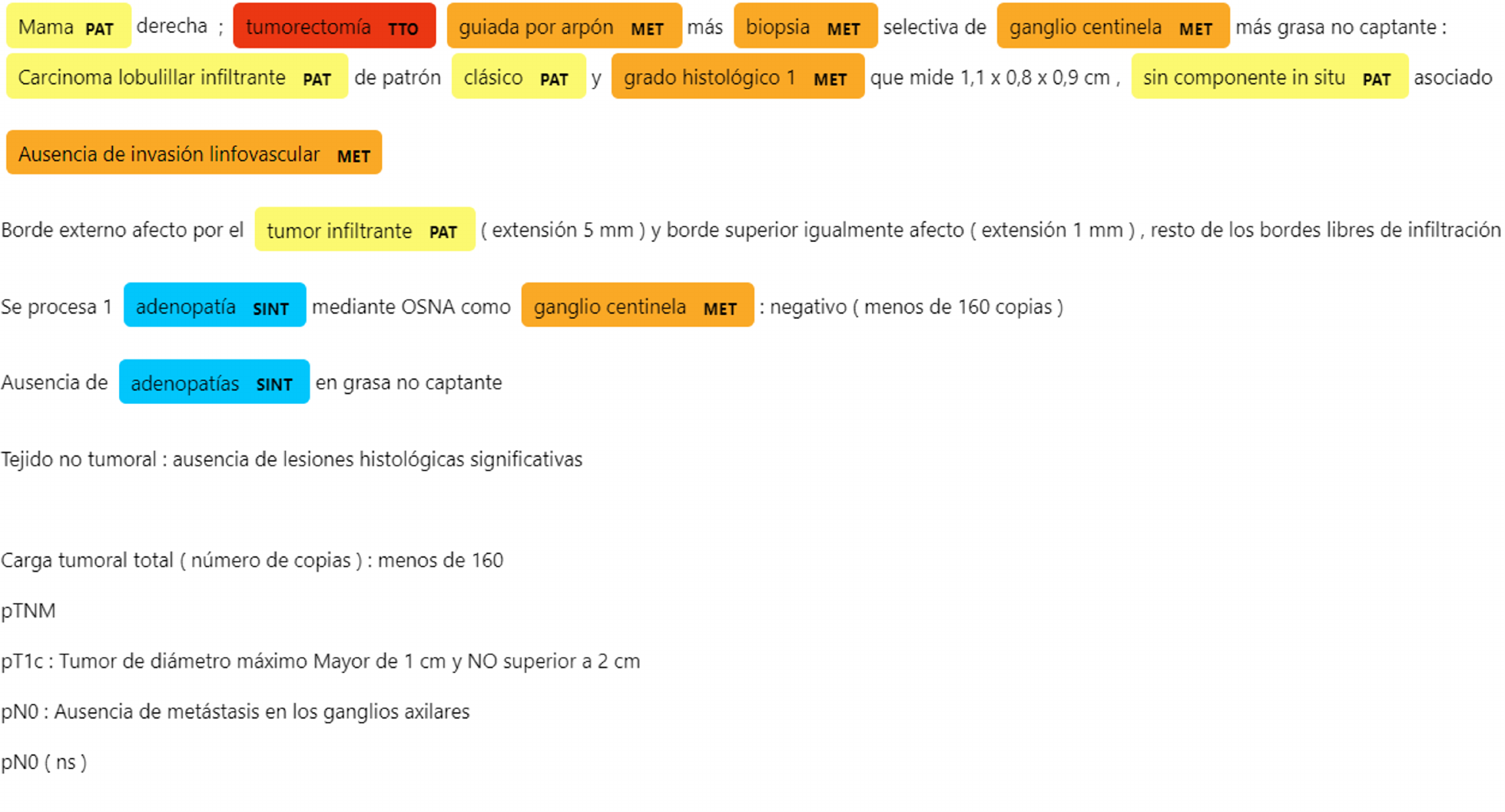}
\caption{Entity extraction from raw text}\label{fig2}
\end{figure}

\begin{figure}[h]
\centering
\includegraphics[width=1.10\textwidth]{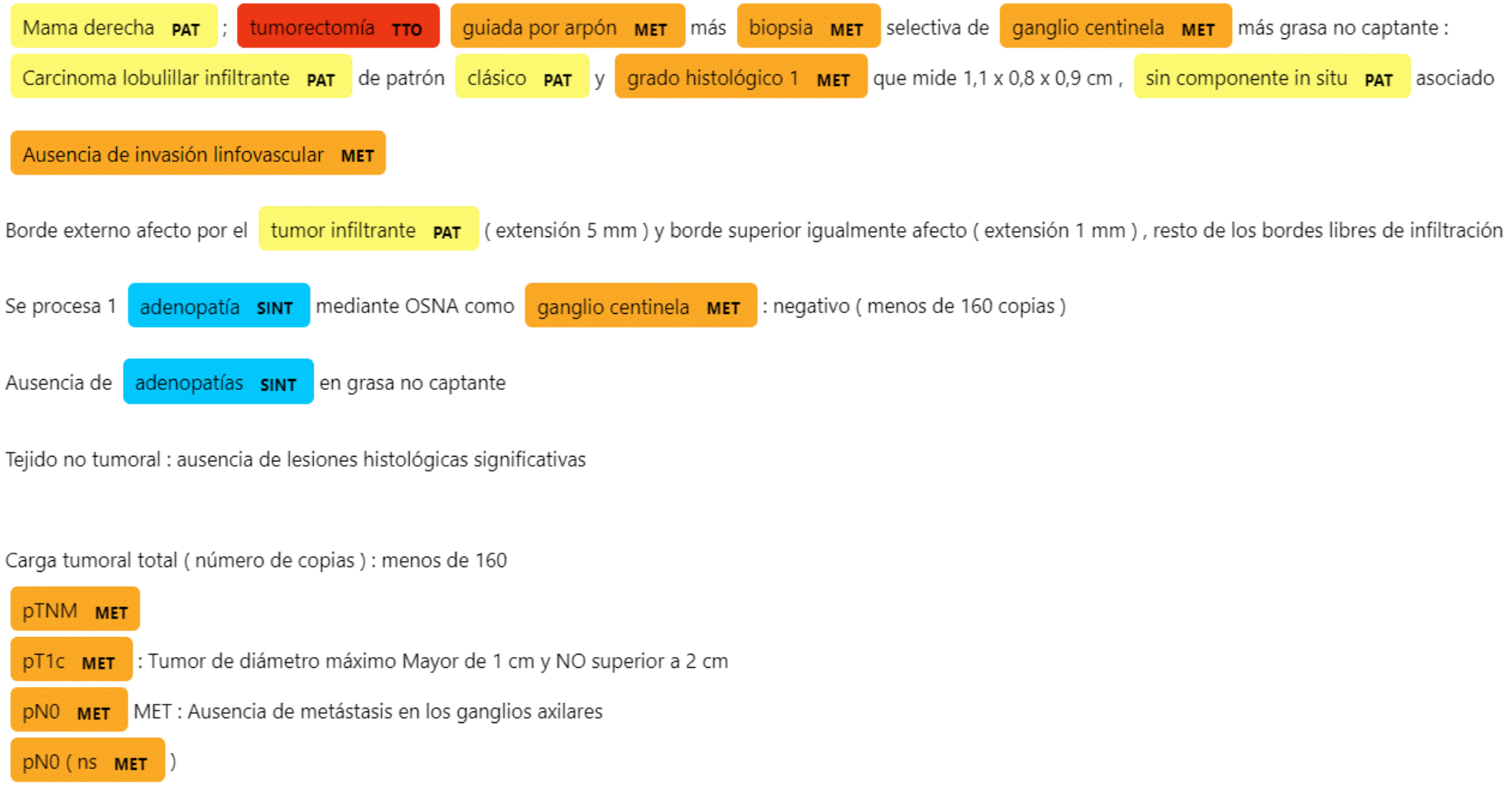}
\caption{Entity extraction from pre-processed text}\label{fig3}
\end{figure}

In this subsection, the results of the manual validation of the breast cancer dataset are presented. The validation process involved the incorporation of a text pre-processing layer using uQuery, GMV's NLP tool, prior to model execution. This pre-processing step was designed to improve the quality of the results and demonstrate the model's robust performance.The impact of this pre-processing layer is illustrated in Figures \ref{fig2} and \ref{fig3}, where the difference in entity detection between unprocessed and optimally pre-processed texts is highlighted.

As shown in \ref{fig2}, when raw text is processed through the NER model, certain entities, such as lung staging (pTNM), are not identified. In contrast, text that has been optimally pre-processed  allows the model to detect all relevant entities, including complete identification of partial entities like \textit{Mama derecha}, which is only partially recognized in unprocessed text. We can see this effect in Figure \ref{fig3}.
This comparison underscores the significant advantage of implementing a pre-processing layer, which maximizes the model's effectiveness and enhances its capability to detect and accurately identify entities. It is also important to note that the pre-processing capabilities of uQuery exceed those discussed in this study, offering superior text cleanliness and a higher detection rate of entities.

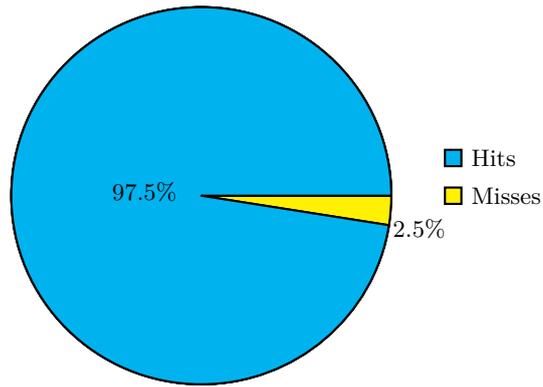
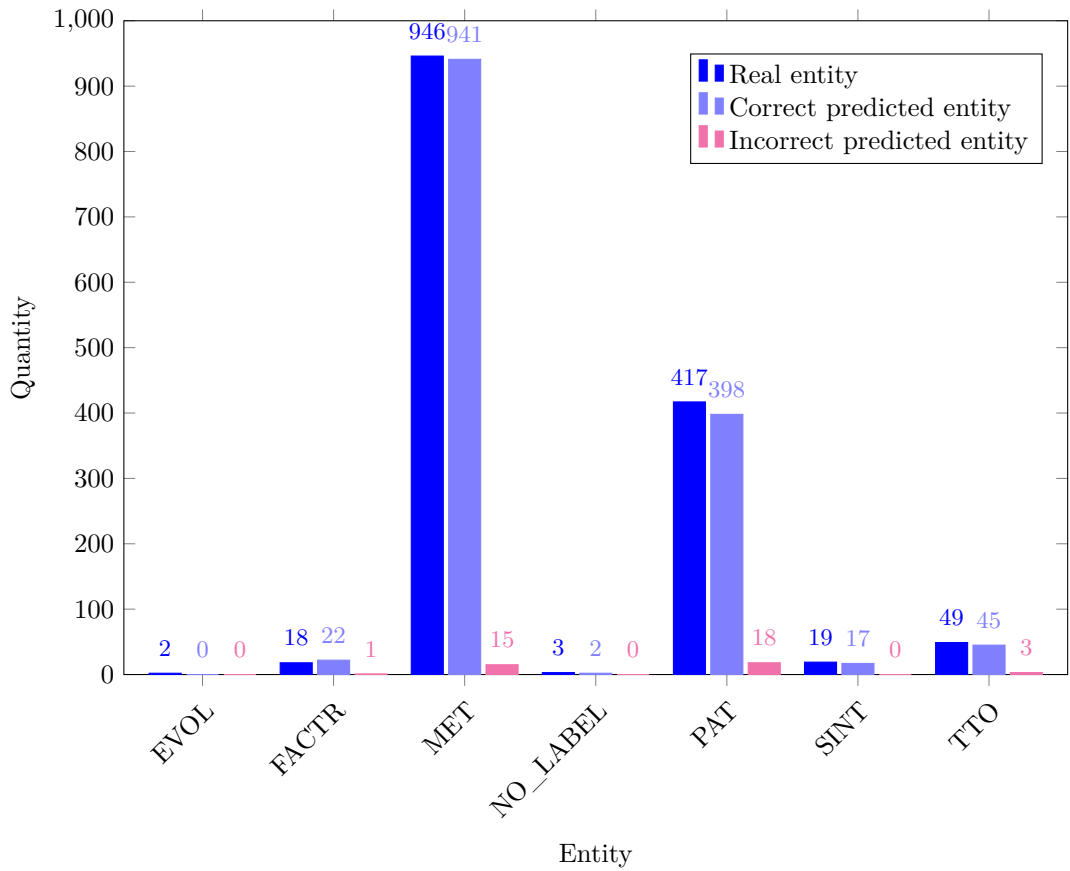
\begin{figure}[!tbp]
  \centering
  \begin{subfigure}[b]{1\textwidth}
    \centering
    \begin{tikzpicture}
    \small
        \pie[
            text=legend, % Muestra la leyenda
            radius=2.5,    % Tamaño del gráfico
            color={cyan!90, yellow}, % Colores para los segmentos
        ]{97.5/Hits, 2.5/Misses} % Datos
        \pie[
            hide number, 
            radius=2.5,    % Tamaño del gráfico
            color={cyan!90, yellow}, % Colores para los segmentos
        ]{97.5/, 2.5/\raisebox{3pt}{$\!\!2.5\%$}} % Datos
        \pie[radius=2.5, color={cyan!90}]{97.5/}
    \end{tikzpicture}
    \caption{}
    \label{fig:f1}
  \end{subfigure}
  \begin{subfigure}[b]{1\textwidth}
    \begin{tikzpicture}
        \begin{axis}[
            ybar, % 
            bar width=12pt,
            width=14cm,
            height=10.25cm,
            enlarge x limits=0.1,
            symbolic x coords={EVOL, FACTR, MET, NO\_LABEL, PAT, SINT, TTO},
            xtick=data,
            xlabel={Entity},
            ylabel={Quantity},
            ymin=0,
            ymax=1000, % Ajuste para la altura del gráfico
            % Colocar la leyenda a la derecha, con las entradas una debajo de otra
            legend style={at={(0.6,0.95)}, anchor=north west, legend cell align={left}},
            xticklabel style={rotate=45, anchor=north east, align=center},
            nodes near coords,
            every node near coord/.append style={font=\small, yshift=3pt}
        ]
    
        % Barras de entidades reales 
        \addplot+[ybar, color=blue!100, fill=blue!100] plot coordinates {
            (EVOL,2)
            (FACTR,18)
            (MET,946)
            (NO\_LABEL,3)
            (PAT,417)
            (SINT,19)
            (TTO,49)
        };
    
        % Barras de predicciones correctas (azul más oscuro) - Apiladas con incorrectas
        \addplot+[ybar, color=blue!50, fill=blue!50] plot coordinates {
            (EVOL,0)
            (FACTR,22)
            (MET,941)
            (NO\_LABEL,2)
            (PAT,398)
            (SINT,17)
            (TTO,45)
        };
    
        % Barras de predicciones incorrectas (rosa claro) - Apiladas sobre las correctas
        \addplot+[ybar, color=magenta!70, fill=magenta!70] plot coordinates {
            (EVOL,0)
            (FACTR,1)
            (MET,15)
            (NO\_LABEL,0)
            (PAT,18)
            (SINT,0)
            (TTO,3)
        };
    
        \legend{Real entity, Correct predicted entity, Incorrect predicted entity}
    
        \end{axis}
    \end{tikzpicture}
    \caption{}
    \label{fig:f1}
  \end{subfigure}
  \caption{(a) Distribution of Hits and Misses in the Validation Set. 
(b) Comparison of Model-Detected Entities with Expert-Annotated Entities: EVOL (Evolution), FACTR (Risk Factors), MET (Method of Diagnosis), NO\_LABEL (Non annotated entities), PAT (Pathology), SINT (Symptomatology), and TTO (Treatment).
}\label{fig4}
\end{figure}

\vspace{3cm}
Following the examination of the methodology, the results are illustrated through graphical representations.
The pie chart in panel (a) of Figure \ref{fig4} provides a visual breakdown of the entity detection results from the validation set. It reveals that 97.5\% of the detected entities are correctly identified, while only 2.5\% are incorrect. This high percentage of accurate detections highlights the model's strong performance in correctly identifying relevant entities.

In addition, a comparative analysis is presented between the entities identified by the model and the true entities as annotated by IIS La Fe experts. The subsequent chart provides a detailed comparison, illustrating the alignment and discrepancies between the model's output and expert annotations. Panel (b) of Figure \ref{fig4} presents this comparative analysis, showing that entities labeled as EVOL are notably absent in the model's detections. Conversely, entities such as MET, PAT, TTO, and SINT show a high degree of alignment with the expert annotations. In some cases, such as MET, the model detected more entities than those actually labeled, which is depicted as an "overshoot" in the chart.

Additionally, an analysis of unlabelled reports reveals that the model's performance is commendable, as it successfully identified relevant entities even in these reports. This highlights the model's robustness and capability in detecting entities beyond the labeled dataset, further demonstrating its effectiveness in practical applications.

These visualizations not only confirm the model's strong performance but also pinpoint areas for potential improvement, providing valuable insights for refining the entity recognition process.

\subsection{Validation results for the complete dataset}\label{subsec5}

The  validation results for the complete dataset, comprising a total of 600 reports, are detailed below. This dataset is categorized as follows:
    \begin{itemize}
        \item 200 reports of Breast Cancer
        \item 400 reports of Lung Cancer
    \end{itemize}

As in the previous section, the dataset was partitioned into training, testing, and validation subsets, following a distribution of 50\% for training, 25\% for testing, and 25\% for validation. Each subset maintains the same proportional representation of report types to ensure consistency across the data splits

As illustrated in Table \ref{tab6}, the labels with the highest representation in the dataset are MET and PAT, consistent with observations from the previous dataset. Conversely, labels such as EVOL and ANTPERSON appear less frequently, reflecting their lower incidence across the complete dataset.

\begin{table}[h]
\caption{Label distribution in the complete dataset}\label{tab6}%
\begin{tabular}{@{}lllllllll@{}}
\toprule
Set & EVOL & FACTR & MUTAC & ANTPERSON & MET & PAT & SINT & TTO\\
\midrule
Train    & 42   & 107  & 242  & 46   & 1916   & 1432 & 327 & 286  \\
Validation   & 21 & 60 & 90  & 17   & 1178  & 746  & 131 &143 \\
Test    & 22 & 20  & 10  & 14  & 711  & 566 & 93 & 105  \\
Complete   & 85 & 187  & 342  & 77  & 3805  & 2744 & 551 & 534  \\

\botrule
\end{tabular}

\end{table}
This comprehensive validation underscores the model's performance across a diverse set of clinical reports, providing a robust assessment of its ability to generalize across different types of cancer and reporting formats. The consistent representation of key labels such as MET and PAT highlights the model's strengths, while the lower frequency of other labels points to areas where further refinement may be needed

\subsubsection{Automatic validation}\label{subsubsec4}

This subsection details the automatic validation results obtained by training the model on the complete dataset. The evaluation metrics used are consistent with those described in Section  \ref{subsubsec2}.

Table \ref{tab7} reveals that the EVOL label shows very low metrics, reflecting its limited representation in the dataset compared to other classes. In contrast, MUTAC achieves the highest metrics across the board. The overall model performance, as summarized in Table \ref{tab8}, indicates a strong performance:

\begin{table}[h]
\caption{Automatic validation results of the complete dataset model per label}\label{tab7}%
\begin{tabular}{@{}lllllllll@{}}
\toprule
Validation Set & EVOL & FACTR & MUTAC & ANTPERSON & MET & PAT & SINT & TTO\\
\midrule
F1          & 0.3684  & 0.6441   & 0.8195 & 0.7027  & 0.7867 & 0.6657 & 0.6593 & 0.7751 \\
Precision   & 0.4118  & 0.6552  & 0.7304  & 0.65    & 0.7809 & 0.6648 & 0.6403 & 0.7619\\
Recall      & 0.3333  & 0.6333  & 0.9333  & 0.7647  & 0.7926  & 0.6666  & 0.6794 & 0.7887\\

\botrule
\end{tabular}

\end{table}

\begin{table}[h]
\caption{Automatic validation results of complete dataset model}\label{tab8}%
\begin{tabular}{@{}llllll@{}}
\toprule
Validation Set & Accuracy & F1 & Precision & Recall & Loss \\
\midrule
Global  & 0.9557  & 0.7411  & 0.7332   & 0.7493 & 0.2565  \\
\botrule
\end{tabular}

\end{table}

The results demonstrate that the model performs exceptionally well, with high accuracy and F1 scores (Table \ref{tab8}). The high recall for MUTAC, which reaches 1, signifies that the model has accurately identified all positive instances of MUTAC in the dataset. However, this high recall for MUTAC may come at the expense of overall accuracy, as the model might be generating more false positives, evidenced by the lower accuracy compared to the validation set. The overall results, shown in Table 8, further confirm that the model's performance on the test set aligns closely with that of the validation set, indicating consistent performance across different subsets of the dataset.

As shown in Table \ref{tab9}, the detailed test results for each label reveal that while the model performs well overall, there are variations in performance across different entity types. Notably, the recall for MUTAC is perfect (1.000), indicating that the model successfully identifies all positive examples for this label. However, this high recall for MUTAC is coupled with a lower accuracy, suggesting the presence of false positives affecting overall accuracy. The overall results, which are very similar to the previous ones, are also shown:

\begin{table}[h]
\caption{Automatic test results of the complete dataset model per label}\label{tab9}%
\begin{tabular}{@{}lllllllll@{}}
\toprule
Test Set & EVOL & FACTR & MUTAC & ANTPERSON & MET & PAT & SINT & TTO\\
\midrule
F1          & 0.375  & 0.56     & 0.8      & 0.8571  & 0.8113 & 0.6564 & 0.6915 & 0.8122 \\
Precision   & 0.3462 & 0.4666   & 0.6666   & 0.8571  & 0.7917 & 0.6754 & 0.6842 & 0.8696\\
Recall      & 0.4091  & 0.7     & 1        & 0.8571  & 0.8320  & 0.6384  & 0.6989 & 0.7619\\

\botrule
\end{tabular}
\end{table}
\begin{table}[h]
\caption{Automatic test results of complete dataset model}\label{tab10}%
\begin{tabular}{@{}llllll@{}}
\toprule
Test Set & Accuracy & F1 & Precision & Recall & Loss \\
\midrule
Global  & 0.9591  & 0.7489  & 0.7435 & 0.7543 & 0.2234  \\
\botrule
\end{tabular}

\end{table}

Table \ref{tab10} summarizes the global test results, showing an accuracy of 95.91\% and an F1-score of 0.7489, consistent with the validation results. The precision and recall values (0.7435 and 0.7543, respectively) further confirm the model's robust performance. The lower loss of 0.2234 supports the conclusion that the model is well-calibrated and generalizes effectively to new data.
In summary, the evaluation across both validation and test datasets demonstrates that the model performs reliably and consistently. The detailed metrics highlight areas where the model excels and where further refinements might be necessary, ensuring its suitability for real-world applications in entity recognition tasks.

\vspace{5cm}
\subsubsection{Manual validation}\label{subsubsec5}

This subsection provides a comprehensive analysis of the manual validation results for the breast and lung cancer datasets, specifically focusing on each of the abovementioned categories. As with the validation described in Section \ref{subsubsec3}, these results were obtained after applying a \textit{text pre-processing layer}. This pre-processing step was crucial for achieving high-quality results and ensuring the model's effective performance.

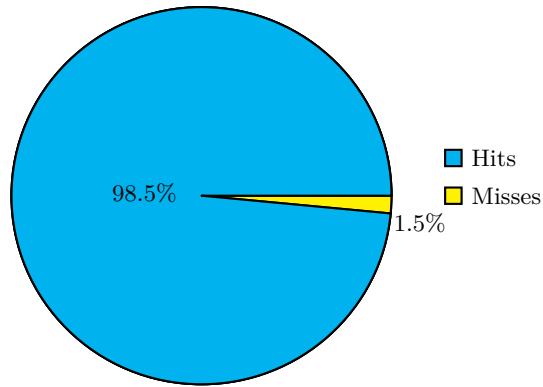
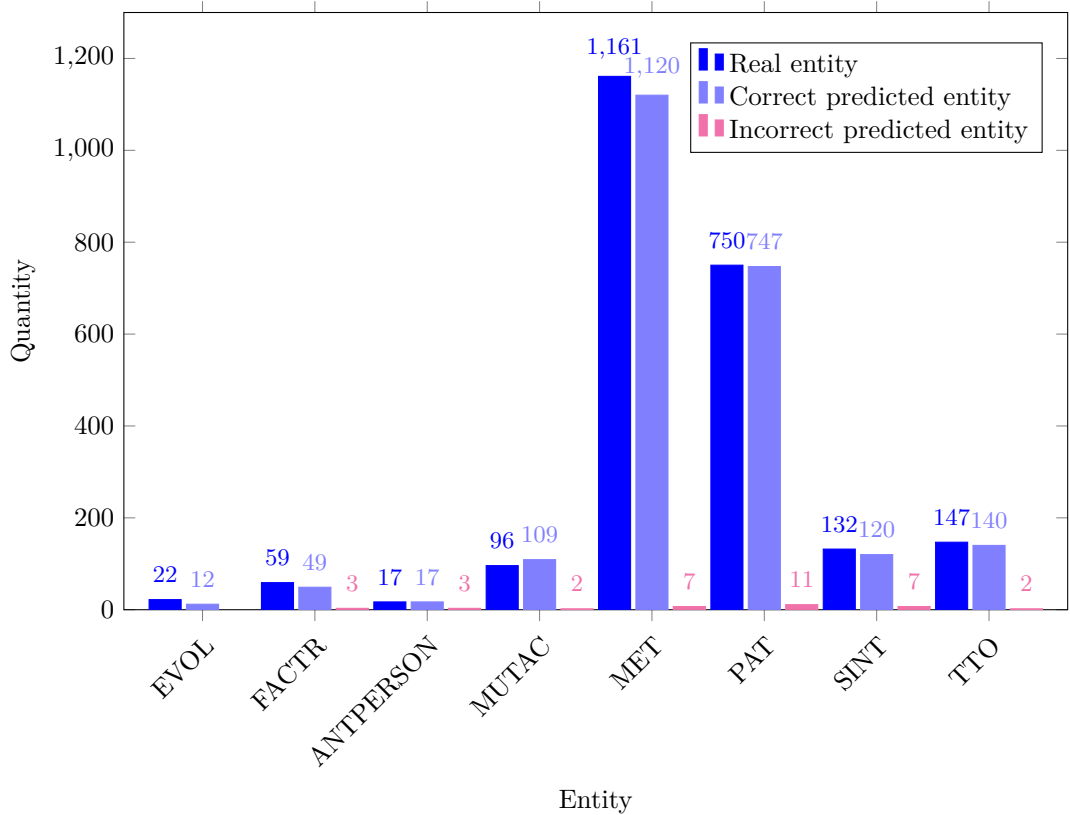
\begin{figure}[!tbp]

  \centering
  \begin{subfigure}[b]{1\textwidth}
  \centering
    \begin{tikzpicture}
    \small
        \pie[
            text=legend, % Muestra la leyenda
            radius=2.5,    % Tamaño del gráfico
            color={cyan!90, yellow}, % Colores para los segmentos
        ]{98.5/Hits, 1.5/Misses} % Datos
        \pie[
            hide number, 
            radius=2.5,    % Tamaño del gráfico
            color={cyan!90, yellow}, % Colores para los segmentos
        ]{98.5/, 1.5/\raisebox{3pt}{$\!\!1.5\%$}} % Datos
        \pie[radius=2.5, color={cyan!90}]{98.5/}
    \end{tikzpicture}
    \caption{}
    \label{fig:f1}
  \end{subfigure}
  \begin{subfigure}[b]{1\textwidth}
  \centering
    \begin{tikzpicture}
        \begin{axis}[
            ybar, % 
            bar width=12pt,
            width=14cm,
            height=9.5cm,
            enlarge x limits=0.1,
            symbolic x coords={EVOL, FACTR, ANTPERSON, MUTAC, MET, PAT, SINT, TTO},
            xtick=data,
            xlabel={Entity},
            ylabel={Quantity},
            ymin=0,
            ymax=1300, % Ajuste para la altura del gráfico
            % Colocar la leyenda a la derecha, con las entradas una debajo de otra
            legend style={at={(0.6,0.95)}, anchor=north west, legend cell align={left}},
            xticklabel style={rotate=45, anchor=north east, align=center},
            nodes near coords,
            every node near coord/.append style={font=\small, yshift=3pt}
        ]
    
        % Barras de entidades reales 
        \addplot+[ybar, color=blue!100, fill=blue!100] plot coordinates {
            (EVOL,22)
            (FACTR,59)
            (ANTPERSON,17)
            (MUTAC,96)
            (MET,1161)
            (PAT,750)
            (SINT,132)
            (TTO,147)
        };

        % Barras de predicciones correctas (azul más oscuro) - Apiladas con incorrectas
        \addplot+[ybar, color=blue!50, fill=blue!50] plot coordinates {
            (EVOL,12)
            (FACTR,49)
            (ANTPERSON,17)
            (MUTAC,109)
            (MET,1120)
            (PAT,747)
            (SINT,120)
            (TTO,140)
        };
    
        % Barras de predicciones incorrectas (rosa claro) - Apiladas sobre las correctas
        \addplot+[ybar, color=magenta!70, fill=magenta!70] plot coordinates {
            (FACTR,3)
            (ANTPERSON,3)
            (MUTAC,2)
            (MET,7)
            (PAT,11)
            (SINT,7)
            (TTO,2)
        };
    
        \legend{Real entity, Correct predicted entity, Incorrect predicted entity}
    
        \end{axis}
    \end{tikzpicture}
    \caption{}
    \label{fig:f1}
  \end{subfigure}
  \caption{(a) Distribution of Hits and Misses in the Validation Set. 
(b) Comparison of Model-Detected Entities with Expert-Annotated Entities: EVOL (Evolution), FACTR (Risk Factors), ANTPERSON (Personal History, specific to lung cancer), MUTAC (Genetic Mutations, specific to lung cancer),MET (Method of Diagnosis), PAT (Pathology), SINT (Symptomatology), and TTO (Treatment).
}\label{fig5}
\end{figure}

Figure \ref{fig5} presents a pie chart illustrating the distribution of hits and misses in the validation set. The chart reveals that an impressive 98.5\% of the identified entities are accurate, with only 1.5\% being incorrect. This high proportion of correct entities underscores the model's robustness and reliability in entity recognition.
Following this, a detailed comparison is provided between the entities identified by the model and those manually annotated (bar chart).

In summary, the manual validation results confirm the model's strong performance, with a very high rate of correct entity identification. The detailed comparison with manual annotations provides valuable insights into the model's accuracy and areas where it may detect additional entities not previously considered, enhancing the overall understanding of its practical application in clinical report analysis.

\vspace{8cm}

\section{Conclusions}\label{sec6}
In conclusion, the evaluation of the NER model on clinical reports for breast and lung cancer has demonstrated its effectiveness in accurately identifying critical clinical entities. The model performed particularly well with entities that were well-represented in the dataset, such as MET and PAT, highlighting its strength in handling commonly occurring clinical data.

However, challenges were encountered with less frequent entities, such as EVOL, indicating areas where the model's performance can be further refined. These difficulties underscore the need for continued improvements, particularly in the detection of underrepresented entities. The integration of a text pre-processing layer proved essential in enhancing the model's accuracy, emphasizing the critical role of pre-processing in optimizing NER performance.

The findings suggest that the model is well-suited for clinical applications, providing robust performance in extracting key information from clinical reports. Nevertheless, further refinement is necessary to improve detection accuracy for less frequent labels, specifically through enhancing the model's sensitivity and specificity.

Overall, this study highlights the potential of fine-tuning NER models to significantly enhance the accuracy and efficiency of information extraction from clinical reports. The insights gained provide a solid foundation for the continued development and application of NER technologies in clinical settings, with promising implications for both research and practical healthcare applications.

\section{List of abbreviations}\label{sec7}
\begin{itemize}
    \item \textbf{NLP: } Natural Language Processing 
    \item \textbf{EHR:} Electronic Health Records 
    \item \textbf{NER:} Named Entity Recognition 
    \item  \textbf{IIS La Fe}: Institute Hospital La Fe 
    \item \textbf{RWE:} Real-World Evidence 
    \item \textbf{EMR:} Electronic Medical Records 
    \item \textbf{CRF:} Conditional Random Fields
    \item \textbf{NSCLC:} Non-Small Cell Lung Cancer 
    \item  \textbf{GCP:} Good Clinical Practice 
    \item \textbf{RoBERTa:} Robustly Optimized BERT Pretraining Approach
\end{itemize}

\bmhead{Data Availability}

The datasets generated and analyzed during the current study are not publicly available due to  lack of consent for public sharing. However, are available from the corresponding author on reasonable request.

\bmhead{Authors' contributions}
J. Moreno-Casanova designed the computational framework, carried out the simulations and analysis and wrote the manuscript.
J.M. Auñon contributed in the design of the solution.
A. Martínez-Pérez and M.E. Pérez-Martínez participated in the identification of the digital patient cohort, the extraction of pseudonymized clinical records, the tagging of the training set, and the writing and review of the manuscript. 
M.E. Gas-López contributed to the writing and review of the manuscript.

\bmhead{Acknowledgements}

This work was supported by grant from R\&D Missions in the Artificial Intelligence program, which is part of the Spain Digital 2025 Agenda and the National Artificial Intelligence Strategy and financed by the European Union through Next Generation EU funds (project TARTAGLIA, exp.MIA.2021.M02.0005). The financial support of GMV is also gratefully acknowledged, along with its continued commitment to research and development.

\bmhead{Authors' information (optional)} Not applicable

\bibliography{sn-bibliography.bib} % common bib file

\end{document}